\documentclass[letterpaper]{article} 
\usepackage{aaai2026}  
\usepackage{times}  
\usepackage{helvet}  
\usepackage{courier}  
\usepackage[hyphens]{url}  
\usepackage{graphicx} 
\usepackage{natbib}  
\usepackage{caption} 
\usepackage{algorithm}
\usepackage{algorithmic}

\usepackage{newfloat}
\usepackage{listings}
\DeclareCaptionStyle{ruled}{labelfont=normalfont,labelsep=colon,strut=off} 
\floatstyle{ruled}
\newfloat{listing}{tb}{lst}{}
\floatname{listing}{Listing}
\usepackage{overpic}
\usepackage{booktabs}
\usepackage{xcolor}
\usepackage{graphicx}
\usepackage{caption}
\usepackage{subcaption}
\usepackage{lipsum} 
\usepackage{multirow}
\usepackage{booktabs} 
\usepackage{amsfonts}
\usepackage{amsmath}
\usepackage{placeins} 
\usepackage{cuted}

\title{{\emph{OmniVDiff}: Omni  Controllable Video Diffusion \\ for Generation and Understanding}}

\author{Dianbing Xi\textsuperscript{\rm 1, \rm 2, \footnotemark[1]}, 
Jiepeng Wang\textsuperscript{\rm 2, \thanks{These authors contributed equally.}, \thanks{These authors served as project leads.}},
Yuanzhi Liang\textsuperscript{\rm 2},
Xi Qiu\textsuperscript{\rm 2},
Yuchi Huo\textsuperscript{\rm 1}, \\
Rui Wang\textsuperscript{\rm 1}\footnotemark[3],
Chi Zhang\textsuperscript{\rm 2}\footnotemark[3],
Xuelong Li\textsuperscript{\rm 2}\thanks{These authors are the corresponding authors.} \\}

\affiliations{
    \textsuperscript{\rm 1}State Key Laboratory of CAD\&CG, Zhejiang University\\
    \textsuperscript{\rm 2}Institute of Artificial Intelligence, China Telecom\\
}

\usepackage{bibentry}

\newcommand{\abbname}{\emph{OmniVDiff}}

\begin{document}

\maketitle
\begin{abstract}
In this paper, we propose a novel framework for controllable video diffusion, \abbname{}, aiming to synthesize and comprehend multiple video visual content in a single diffusion model. 
To achieve this, \abbname{} treats all video visual modalities in the color space to learn a joint distribution, while employing an adaptive control strategy that dynamically adjusts the role of each visual modality during the diffusion process, either as a generation modality or a conditioning modality.
Our framework supports three key capabilities: (1) {Text-conditioned video generation}, where all modalities are jointly synthesized from a textual prompt; (2) {Video understanding}, where structural modalities are predicted from rgb inputs in a coherent manner; and (3) {X-conditioned video generation}, where video synthesis is guided by fine-grained inputs such as depth, canny and segmentation. Extensive experiments demonstrate that \abbname{} achieves state-of-the-art performance in video generation tasks and competitive results in video understanding. 
Its flexibility and scalability make it well-suited for downstream applications such as video-to-video translation, modality adaptation for visual tasks, and scene reconstruction. Our project page: \texttt{{https://tele-ai.github.io/OmniVDiff/}}.
\end{abstract}

\begin{figure}[t]

		\centering
        \begin{overpic}[width=\linewidth]{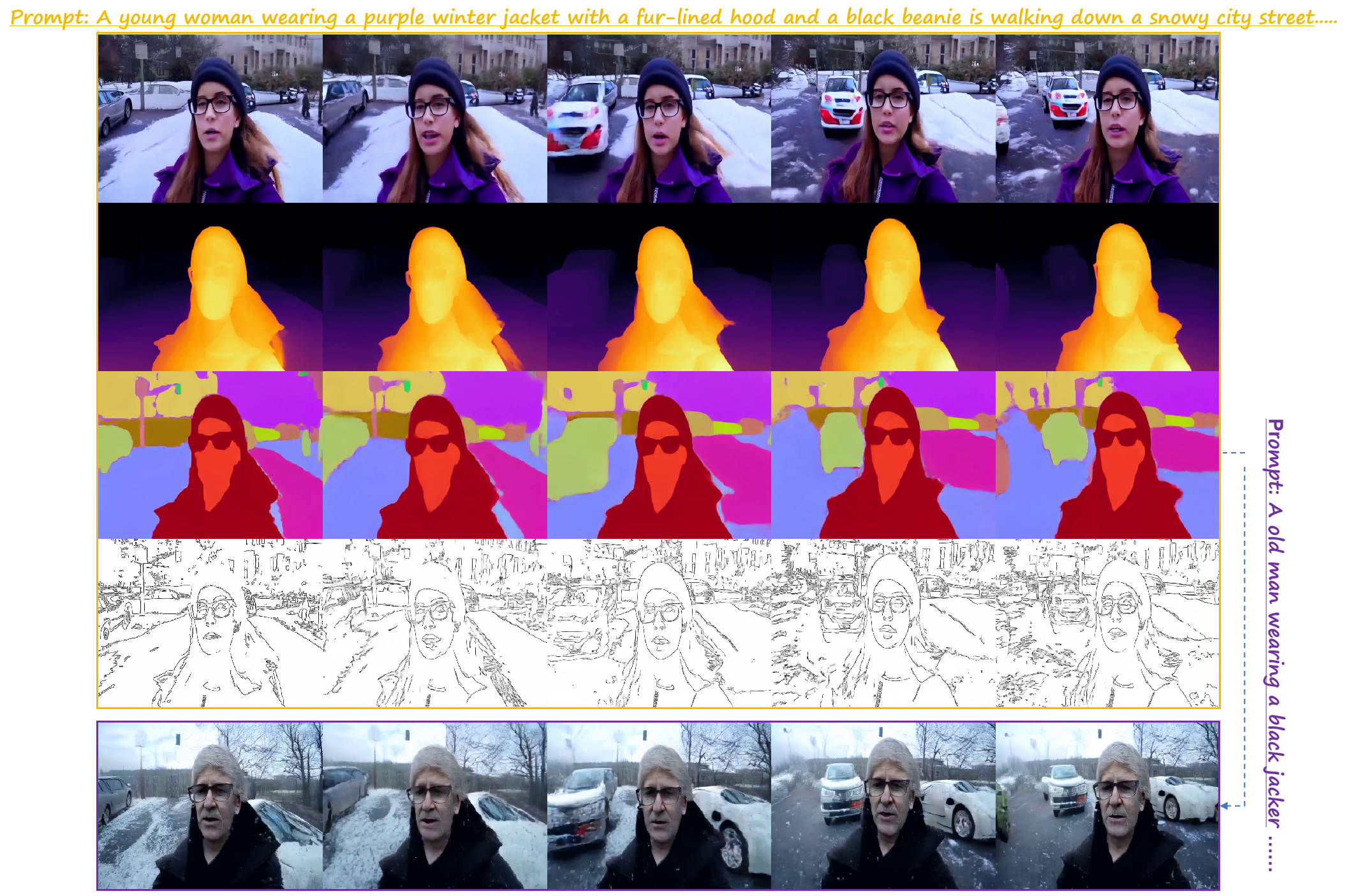}
              \put(-3,0){%
                \rotatebox{90}{%
                  \scalebox{0.85}{\textbf{(a) Text to multi-modal video generation}}%
                }%
              }
              \put(3.5,16){%
                \rotatebox{90}{%
                  \scalebox{0.8}{canny}%
                }%
              }
              \put(3.5,29){%
                \rotatebox{90}{%
                  \scalebox{0.8}{seg}%
                }%
              }
             \put(3.0,42){%
                \rotatebox{90}{%
                  \scalebox{0.8}{depth}%
                }%
              }
              
              \put(3.0,56){%
                \rotatebox{90}{%
                  \scalebox{0.8}{rgb}%
                }%
              }

            \put(95,60){%
              \rotatebox{-90}{%
                \scalebox{0.80}{\textbf{(b) Seg‐conditioned video generation}}%
              }%
            }
           \put(89.5,38){%
            \rotatebox{-90}{%
              \scalebox{0.8}{seg}%
            }%
          }
          
          \put(89.5,5){%
            \rotatebox{-90}{%
              \scalebox{0.8}{rgb}%
            }%
          }

        \end{overpic}

        \captionof{figure}{\textbf{Omni controllable video generation and understanding.} 
        Given a text prompt, (a) \abbname{} generates high-quality rgb videos while simultaneously producing aligned multi-modal visual understanding outputs (i.e., depth, segmentation and canny). 
        Additionally, (b) \abbname{} supports X-conditioned video generation within a unified framework, such as seg-conditioned video generation.
        }
		\label{fig:teaser}
\end{figure}

\section{Introduction}
\label{sec:intro}

Diffusion models have achieved remarkable progress in image~\cite{rombach2022high} and video generation~\cite{blattmann2023svd,kong2024hunyuanvideo,yang2024cogvideox}, demonstrating strong controllability and generalization through large-scale training. 
For controllable video generation, models typically employ conditions such as depth~\cite{guo2024sparsectrl,liu2024stablev2v,xing2024make}, segmentation~\cite{zhao2023make,khachatryan2023text2video,hu2025animate}, or canny edges~\cite{lv2024gpt4motion} to guide the diffusion process. 
By fine-tuning pretrained text-to-video (T2V) models~\cite{blattmann2023svd,yang2024cogvideox}, these approaches achieve high-quality controllable generation. 
However, most existing methods rely on task-specific fine-tuning and external expert models to obtain conditional modalities, which limits scalability and increases computational cost. 
Recent works further explore joint multi-modal generation~\cite{zhai2024idol,VideoJAM,byung2025jointdit,wang2025mmgen,vace,huang2025voyagerlongrangeworldconsistentvideo}, yet they primarily focus on joint synthesis and lack support for generative understanding or conditional control. 
Overall, while video diffusion models show strong potential, their limited adaptability remains a key obstacle to developing a unified and efficient framework for diverse video-related tasks.

Recently, several concurrent studies in the image domain explored unifying multiple tasks within a single diffusion framework, by treating image-level tasks as a sequence of image views \cite{le2024onediff,chen2024UniReal, wang2025mmgen,zhao2025diception} (analogous to video generation). For example, the depth-conditioned generation can be regarded as a two-view (depth and rgb) diffusion task.
While this approach has been effective for image-based tasks, extending it to video generation presents significant challenges. Unlike images, videos introduce an additional temporal dimension.
Treating modalities as distinct video sequences would significantly increase the token length and computation cost in the transformer-based diffusion process, especially considering the quadratic computational complexity in the attention mechanism \cite{vaswani2017attention}.
The challenge of extending such approaches into a unified video diffusion framework that can handle both conditioned and unconditioned generation remains largely unexplored.

In this work, we propose \abbname{}, a unified framework for controllable video generation. 
Our approach comprises two key components: (1) a multi-modal video diffusion architecture and (2) an adaptive modality control strategy, jointly enabling efficient handling of diverse visual modalities for both generation and understanding. 
(1) In the diffusion network, we extend the input noise dimensionality to match the number of modalities, allowing the model to process multiple visual inputs seamlessly. Distinct projection heads generate modality-specific outputs while preserving a unified framework. 
(2) To enhance adaptability, we introduce a flexible control strategy that dynamically assigns each modality as generative or conditional. For generative modalities, inputs are blended with noise, while conditional ones retain their original signals. This distinction is reinforced through learnable modality-specific embeddings. 
Through this design, our method achieves fine-grained control across modalities, providing a unified and adaptable framework for video generation and understanding tasks.

To this end, we focus on four representative visual modalities: rgb, depth, segmentation, and canny. To train our unified diffusion model, we construct a paired multi-modal dataset by filtering a subset of videos from Koala-36M~\cite{wang2024koala} and applying expert models to generate high-quality pseudo-labels for each modality. 

We evaluate our approach on a broad range of tasks, including text-to-video generation, X-conditioned video generation, and multi-modal video understanding, and further assess its generalization to downstream tasks such as video-to-video style transfer and super-resolution. Extensive experiments demonstrate the robustness and versatility of our unified framework.

In summary, our main contributions are as follows:

\begin{itemize}
    \item A unified controllable diffusion framework, supporting text-conditioned video generation, controllable generation with structural modalities (depth, canny, segmentation), and video understanding within a single model.
    
    \item An adaptive modality control strategy that dynamically determines the role of each modality (generation or conditioning), enabling fine-grained control and enhancing task adaptability.

    \item Comprehensive evaluation across generation and understanding tasks, demonstrating controllable video generation without expert dependency, and generalization to applications such as style transfer and super-resolution.

\end{itemize}

\section{Related Works}
\label{sec:related_works}

\begin{figure*}[t]
  \centering
    \begin{overpic}[width=\linewidth]{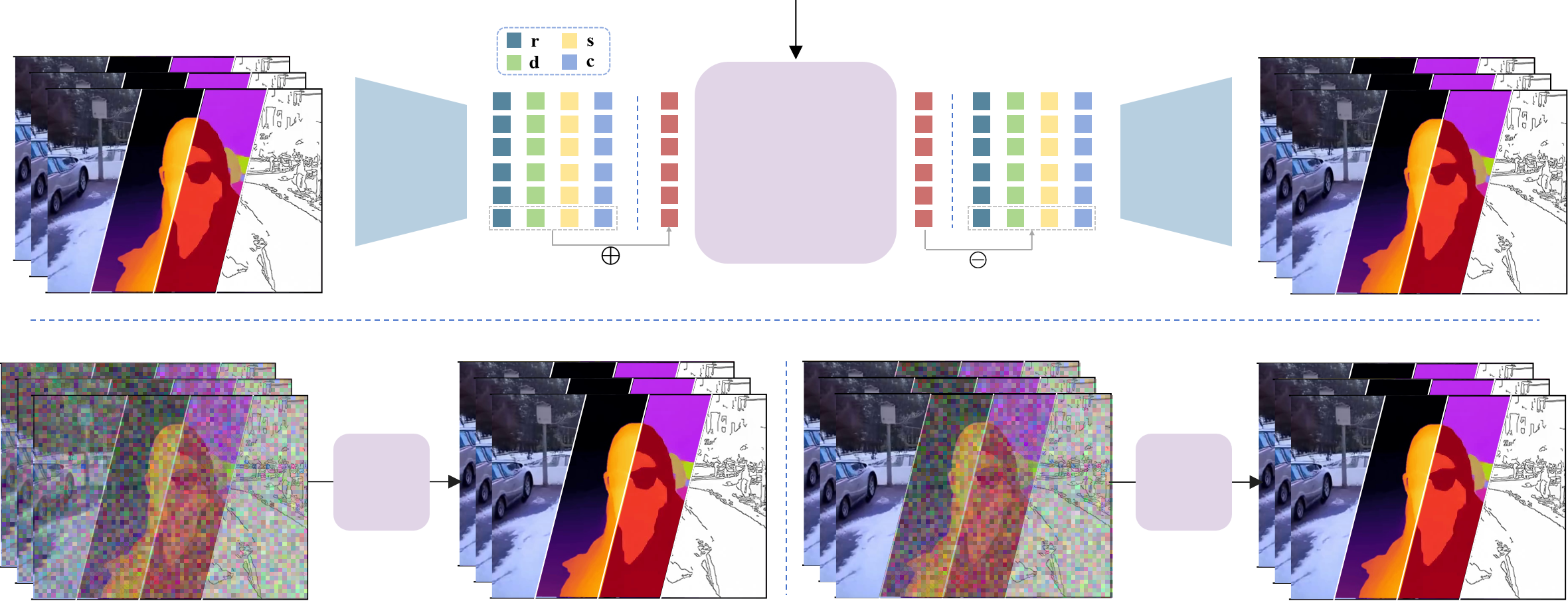}
        \put(4,36){\small rgb}
        \put(7,36){\small depth}
        \put(12,36){\small seg}
        \put(16,36){\small canny}

        \put(26,27){\large $\mathcal{E}$ }
        \put(74,27){\large $\mathcal{D}$ }
        \put(41.5,36.5){\small Text prompt }
        
        \put(46,27){ \abbname }
        
        \put(21.7,7.1){\small \emph{OVDiff} }
        \put(72.2,7.1){ \small \emph{OVDiff} }

        \put(10, 39){\small (a) Video encoding}
        \put(45, 39){\small (b) \abbname}
        \put(75, 39){\small (c) Video decoding}
        
        \put(35, 21){\small $e_m$}
        
        \put(42, 21){\small $x_i$}
        \put(58, 21){\small $x_o$}

        \put(10, -1.5){\small (d) Multi-modal video generation}
        \put(60, -1.5){\small  (e) X-conditioned generation/understanding}

        \put(3,16){\small [rgb}
        \put(7,16){\small depth}
        \put(12,16){\small seg}
        \put(16,16){\small canny] + noise}

        \put(31,16){\small rgb}
        \put(35,16){\small depth}
        \put(40,16){\small seg}
        \put(44,16){\small canny}

        \put(54,16){\small \textbf{\textcolor{blue}{rgb}}}
        \put(57,16){\small [depth}
        \put(63,16){\small seg}
        \put(66,16){\small canny] + noise}

        \put(82,16){\small \textcolor{blue}{rgb}}
        \put(86,16){\small depth}
        \put(91,16){\small seg}
        \put(95,16){\small canny}

    \end{overpic}
    \caption{\textbf{Method overview}.
    (a) Given a video with four paired modalities, we first encode it into latents using a shared 3D-VAE encoder; (b) Then, concatenate them along the channel dimension and apply noise for video diffusion, where the denoised latents are then decoded into their respective modalities via modality-specific decoding heads; (c) Finally, each modality can be reconstructed into color space by the 3D-VAE decoder.
    During inference, the model enables various tasks by dynamically adjusting the role of each modality: (d) Text-to-video generation, where all modalities are denoised from pure noise, and  (e) X-conditioned generation, where the condition X is given and other modalities are denoised from pure noise. If X is rgb modality, the model will perform generative understanding.
   }
   \label{fig:method_overview}
\end{figure*}

\subsection{Text-to-video Diffusion}
Text-to-video (T2V) diffusion models have made significant progress in generating realistic and temporally consistent videos from text prompts \cite{kong2024hunyuanvideo,polyak2025moviegencastmedia}. 
SVD \cite{blattmann2023svd}, VDM \cite{ho2022video} and following works \cite{hong2022cogvideo} explore extending image diffusion models \cite{rombach2022high} for video synthesis with spatial and temporal attention 
\cite{chen2024videocrafter2,feng2024ccedit}. 
Recent methods also introduce 3D Variational Autoencoder (VAE) to compress videos across spatial and temporal dimensions, improving compression efficiency and video quality\cite{yang2024cogvideox,kong2024hunyuanvideo,wan2025}.
However, these approaches primarily focus on text-conditioned video generation and lack fine-grained control over video attributes. Tasks such as depth-guided or segmentation-conditioned video generation remain challenging, as text-to-video diffusion models do not explicitly support these controls. Meanwhile, all these methods mainly focus on the rgb modality output, without considering the generative capability of other visual modalities.

\subsection{Controllable Video Diffusion}
To address controllable video generation, many methods try to introduce additional conditioning signals to guide the diffusion process. 
Depth maps can provide accurate geometric and structural information, ensuring realistic spatial consistency across frames \cite{xing2024make,chen2023control,zhang2023controlvideo}. Pose conditioning ensures accurate human motion synthesis by constraining body articulation and joint movements\cite{gan2025humandit,hu2025animate}. Optical flow constrains motion trajectories by capturing temporal coherence and movement patterns, enhancing dynamic realism \cite{liu2024stablev2v}.
However, these existing methods face two major challenges: (1) Fine-tuning for each task: incorporating new control signals typically requires task-specific fine-tuning on large-scale diffusion architectures, making these models computationally expensive and difficult to scale across diverse control modalities. (2) Dependency on external expert models: most approaches rely on pre-extracted conditioning signals from external expert models. For example, in depth-conditioned video generation, a separate depth estimation model is first applied to a reference video, and the estimated depth is then fed into a distinct video diffusion model for generation. This results in a multi-step, non-end-to-end pipeline where each component is trained separately, potentially causing inconsistencies across models and complex operations.

\subsection{Unified Multi-modal Video Generation}
Some efforts have attempted to unify multi-modal generation within a single diffusion model \cite{zhai2024idol,wang2024lavin,VideoJAM,byung2025jointdit,wang2025mmgen,vace,huang2025voyagerlongrangeworldconsistentvideo}.
VideoJAM~\cite{VideoJAM} jointly forecasts rgb frames and optical flow. However, such approaches primarily focus on 
joint modeling of two modalities, offering limited support for conditional generation and understanding.
In addition, DiffusionRenderer~\cite{liang2025diffusionrenderer} addresses both inverse and forward rendering, but relies on two separate models, where the forward rendering process is treated as conditional generation. 
Similarly, UDPDiff~\cite{yang2025unifieddensepredictionvideo} supports joint generation of RGB with either depth or segmentation, 
yet it cannot synthesize all three modalities simultaneously or perform video understanding within a unified framework.
Concurrently, Aether~\cite{aetherteam2025aethergeometricawareunifiedworld} proposes a unified framework that supports both video understanding and joint multi-modal generation across rgb, depth, and camera pose. However, its primary focus lies in geometric world modeling, while generalization to a wider range of modalities like semantic masks and enabling flexible modality-conditioned controllable generation and understanding remains largely under-explored.
In this paper, our method addresses these challenges by introducing a unified framework that allows fine-grained adaptive modality control. 
Unlike prior works, we do not require separate fine-tuning for each control modality and eliminate the reliance on external expert models by integrating multi-modal understanding and generation into a single pipeline. 
This enables more efficient, end-to-end controllable video synthesis, significantly improving scalability and coherence across video generation tasks.

In this work, we address these challenges by introducing a unified framework that enables fine-grained, adaptive modality control. 
Unlike prior approaches, our method eliminates the need for per-modality fine-tuning and external expert models, integrating multi-modal understanding and generation into a single end-to-end pipeline. 
This design facilitates efficient and coherent controllable video synthesis, improving both scalability and consistency across tasks.

\section{Method}
\label{sec:methon}

In this section, we introduce \abbname, a unified framework for video generation and understanding, extending video diffusion models to support multi-modal video synthesis and analysis. We begin with a preliminary introduction to video diffusion models.
Then, we detail our network design and adaptive control strategy, which enable seamless handling of text-to-video generation, modality-conditioned video generation, and multi-modal video understanding.
Finally, we describe our training strategy.
Figure \ref{fig:method_overview} provides an overview of our framework.

\subsection{Preliminary}\label{sec:preliminary}
Video diffusion models generate videos by progressively refining noisy inputs through a denoising process, following a learned data distribution. CogVideoX \cite{yang2024cogvideox}, one of the state-of-the-art text-to-video diffusion models, incorporates a 3D Variational Autoencoder (3D-VAE) to efficiently compress video data along both spatial and temporal dimensions, significantly reducing computational costs while preserving motion consistency.  

Given an input video \(V\in \mathbb{R}^{f \times h \times w \times c}\), where \(f, h, w, c\) denote the number of frames, height, width, and channels, respectively, the 3D-VAE encoder downsamples it using a spatiotemporal downsampling factor of (8,8,4) along the height, width, and frame dimensions: 
\(
F=\frac{f}{4}, \quad H=\frac{h}{8}, \quad W=\frac{w}{8}
\). This process captures both appearance and motion features while significantly reducing the memory and computational requirements of the diffusion process.  
The video diffusion model operates in this latent space, iteratively denoising \(\mathbf{x}_t\) through a learned reverse process. The training objective minimizes the mean squared error (MSE) loss for noise prediction:  
\begin{equation}
\mathcal{L}_{\text{denoise}} = \mathbb{E}_{\mathbf{x}_0, t, \epsilon} \left[ \|\epsilon - \epsilon_{\theta}(\mathbf{x}_t, t)\|^2 \right]
\end{equation}
where \(\epsilon_{\theta}\) is the noise prediction model, \(\mathbf{x}_t\) is the noisy latent at timestep \(t\), and \(\epsilon\) is the added noise.

\subsection{Omni Video Diffusion}\label{sec:network}

\paragraph{Multi-modal video diffusion architecture}
To achieve omni-controllable video diffusion, we design a novel video diffusion architecture that learns a joint distribution over multiple visual modalities. Building upon the pretrained text-to-video diffusion model CogVideoX, we extend the input space to accommodate multiple modalities. On the output side, we introduce \textbf{modality-specific projection heads(MSPH)} to recover each modality separately. 
This design enables our architecture to seamlessly support multi-modal inputs and outputs, ensuring flexible and controllable video generation.

Given a video sequence and its paired visual modalities \(V = \{V_r, V_d, V_s, V_e\}\), where \(V_r\), \(V_d\), \(V_s\), and \(V_e\) represent rgb, depth, segmentation, and canny, respectively, we first encode them into a latent space using a pretrained 3D-causal VAE encoder \( \mathcal{E}\) \cite{yang2024cogvideox}. Each modality is mapped to latent patches to get the noisy latents:  
\begin{equation}
x_{m} = \mathcal{E}(V_m), \quad m \in \{r, d, s, c\}.    
\end{equation}
where $x_m \in \mathbb{R}^{F \times H \times W \times C}$ and \(F, H, W, C\) denote the number of frames, height, width, and latent channels, respectively.  

Next, we blend the latent representations of each modality with noise:
$$
x_{m}^t = (1 - t) \cdot \epsilon + t \cdot x_{m}.
$$


The noisy latents are then concatenated along the channel dimension to form a unified multi-modal representation:
$
x_i = \text{Concat}(x_{r}^t, x_{d}^t, x_{s}^t, x_{c}^t).
$
This fused representation serves as the input to the diffusion transformer, enabling the video diffusion model to learn a joint distribution over the multiple modalities.

On the output side, we employ modality-specific projection heads \(H_m\), where each head is responsible for reconstructing the noise output \(\epsilon_m\) of a specific modality from the diffusion transformer output $x_o$:  
\begin{equation}
\epsilon_m = H_m(x_o)
\end{equation}

Specifically, we adopt the original rgb projection head from CogVideoX and replicate it for each modality, rather than simply extending the output channels of a shared rgb head. This design better accommodates the distinct characteristics of different modalities. 
Finally, the denoised latents are decoded back into the color space using the pretrained 3D-VAE decoder $\mathcal{D}$ \cite{yang2024cogvideox}, producing high-fidelity multi-modal video outputs.

\paragraph{Adaptive modality control strategy} 
A key challenge in unified video generation is determining the role of each modality—whether it serves as a generation signal or a conditioning input. To address this, we introduce an \textbf{adaptive modality control strategy(AMCS)} that dynamically assigns roles to different modalities based on the task.  

During training, generation modalities are blended with noise before being fed into the diffusion model, while conditioning modalities remain unchanged and are concatenated with the noisy inputs of other modalities to serve as conditioning signals. This mechanism ensures flexible and adaptive control over different modalities, allowing the model to seamlessly handle diverse tasks within a unified framework.
Specifically, in a text-to-video generation task, all modalities are generated from pure noise, meaning they act as generation signals. In an \(X\)-conditioned generation task, where \(X \) represents depth, segmentation, or canny, the conditioning modality \(X\) is provided as input directly without blending with noise and concatenated with the noisy latent representations of other modalities. Notably, if $X$ represents the rgb modality, the model instead performs a video understanding task and predicts corresponding multi-modal outputs.  
\begin{equation}
\mathbf{x}_{m}^t =
\begin{cases} 
(1 - t) \cdot \epsilon + t \cdot x_{m}, & \text{if } m \text{ is for generation} \\
x_{m}, & \text{if } m \text{ is for conditioning}
\end{cases}    
\end{equation}

To further enhance the diffusion model's ability to distinguish modality roles, we introduce a modality embedding \( \mathbf{e}_m \) that differentiates between generation (\( \mathbf{e}_g \)) and conditioning (\( \mathbf{e}_c \)) roles, which can be directly added to the diffusion model input $\mathbf{x}_{m}^t$.  
\begin{equation}
\mathbf{e}_m =
\begin{cases} 
\mathbf{e}_g, & \text{if } m \text{ is for generation} \\
\mathbf{e}_c, & \text{if } m \text{ is for conditioning}
\end{cases}    
\end{equation}

\begin{equation}
\mathbf{x}_{m}^{t,'} = \mathbf{x}_{m}^t + \mathbf{e}_m
\end{equation}
This strategy enables flexible and efficient control, allowing the model to seamlessly adapt to different tasks without requiring separate architectures for each modality.  

 

\begin{table*}[t]
\centering
\resizebox{\textwidth}{!}{
\begin{tabular}{lccccccc}
\toprule
 & subject consistency & b.g. consistency & motion smoothness & dynamic degree & aesthetic quality & imaging quality & weighted average \\
\midrule
CogVideoX\cite{yang2024cogvideox} & 95.68 & 96.00 & 98.21 & \textbf{53.98} & 50.75 & 65.77 & 72.25 \\
\abbname(ours) & \textbf{97.78} & \textbf{96.26} & \textbf{99.21} & 49.69 & \textbf{51.47} & \textbf{67.13} & \textbf{72.78}  \\

\bottomrule
\end{tabular}}
\caption{\textbf{VBench metrics for text-conditioned video generation.} We compare our method, \abbname, with prior baseline CogVideoX. For each metric group, the best performance is shown in \textbf{bold}.
}
\label{tab:vbench_prediction}
\end{table*}

\begin{table*}[t]
\centering
\resizebox{\textwidth}{!}{
\begin{tabular}{lccccccc}
\toprule
Model & subject consistency & b.g. consistency & motion smoothness & dynamic degree & aesthetic quality & imaging quality & weighted average \\
\midrule

\multicolumn{8}{l}{\textit{text+depth}} \\
\hspace{1em}Control-A-Video\cite{chen2023control} & 89.99 & 91.63 & 91.90 & 40.62 & 48.67 & 68.69 & 68.53 \\
\hspace{1em}ControlVideo\cite{zhang2023controlvideo} & 95.50 & 94.17 & 97.80 & 18.35 & \textbf{57.56} & \underline{70.09} & 70.71 \\
\hspace{1em}Make-your-video\cite{xing2024make} & 90.04 & 92.48 & 97.64 & \underline{51.95} & 44.67 & \textbf{70.26} & 70.17 \\
\hspace{1em}VideoX-Fun\cite{VideoXFun2024} & \underline{96.25} & \underline{95.73} & \underline{98.90} & 50.43 & \underline{55.81} & 55.38 & \underline{72.85} \\
\hspace{1em}\abbname(ours) & \textbf{97.96} & \textbf{96.66} & \textbf{99.18} & \textbf{53.32} & 52.95 & 67.26 & \textbf{73.45} \\

\midrule
\multicolumn{8}{l}{\textit{text+canny}} \\
\hspace{1em}CogVideoX+CTRL\cite{cogvideoxControlNet2024} & 96.26 & 94.53 & 98.42 & \underline{53.44} & 49.34 & 55.56 & 70.13 \\
\hspace{1em}Control-A-Video\cite{chen2023control} & 89.81 & 91.27 & 97.86 & 41.79 & 47.23 & \textbf{68.77} & 69.31 \\
\hspace{1em}ControlVideo\cite{zhang2023controlvideo} & 95.23 & 94.00 & 97.12 & 17.58 & \textbf{55.81} & 55.38 & 67.72 \\
\hspace{1em}VideoX-Fun\cite{VideoXFun2024} & \underline{96.69} & \underline{95.41} & \underline{99.15} & 50.78 & \underline{52.99} & 66.76 & \underline{72.73} \\
\hspace{1em}\abbname(ours) & \textbf{97.84} & \textbf{95.55} & \textbf{99.23} & \textbf{53.53} & 52.34 & \underline{67.14} & \textbf{73.14} \\

\midrule
\multicolumn{8}{l}{\textit{text+segment}} \\
\hspace{1em}\abbname(ours) & \textbf{97.97} & \textbf{95.81} & \textbf{99.31} & \textbf{53.18} & \textbf{53.37} & \textbf{67.51} & \textbf{73.42} \\

\bottomrule
\end{tabular}
}
\caption{\textbf{VBench metrics for depth-, canny-, and segmentation-conditioned video generation.} For each condition type, the best performance is shown in \textbf{bold}, and the second-best is marked with an \underline{underline}.}
\label{tab:comp_c2v_vbench}
\end{table*}


\subsection{Training}\label{sec:training}

\paragraph{Training data}
Training a unified multi-modal model requires a large amount of paired data across modalities such as segmentation and depth. However, high-quality labeled video datasets are inherently scarce, posing a significant bottleneck. To address this, we employ expert models to generate pseudo labels for unlabeled videos, allowing us to efficiently construct a large-scale multi-modal dataset without manual annotation. Benefiting from the rapid advancements of 2D foundation models \cite{sam2,video_depth_anything}, these expert models can provide high-quality annotations at scale, enabling us to leverage large volumes of raw video data for effective training.
Specifically, for video depth, we use Video Depth Anything~\cite{video_depth_anything} to generate temporally consistent depth maps across video sequences. For segmentation, we apply Semantic-SAM~\cite{li2023semanticsam} on the first frame for instance segmentation, then propagate the results to subsequent frames using SAM2~\cite{sam2} to maintain semantic consistency. For canny edges, we adopt the OpenCV implementation of the Canny algorithm~\cite{canny1986computational} for edge detection.

In total, we processed 400K video samples, randomly sampled from the Koala-36M~\cite{wang2024koala} dataset. 
The inference of the video depth estimation model took approximately 3 days, while the video segmentation model required around 5 days, both conducted using 8 NVIDIA H100 GPUs in parallel. 

\paragraph{Training loss}
We optimize our unified video generation and understanding framework using a multi-modality diffusion loss, ensuring high-quality generation while maintaining flexibility across different modalities. 
For each modality, we apply an independent denoising loss. 
If a modality serves as a conditioning input, the denoising loss is skipped for that modality, ensuring it only guides the generation process without being explicitly optimized. 
The final objective is:  
\begin{equation}
\mathcal{L} = \sum_{m, m \notin Cond}
 \mathbb{E}_{\mathbf{x}_m, t, \epsilon, m} \left[ \|\epsilon - \epsilon_{\theta}(\mathbf{x}_{m}^{t,'}, t, e_m)\|^2 \right]
\end{equation}

This approach provides adaptive supervision, enabling flexible role assignments for modalities and allowing the model to seamlessly transition between generation and conditioning tasks.

\section{Experiments}
\label{sec:exp}

\subsection{Implementation Details}
We fine-tune our model based on CogVideoX \cite{yang2024cogvideox}, a large-scale text-to-video diffusion model. Specifically, we adopt CogVideoX1.5-5B as the base model for our fine-tuning. The fine-tuning process follows a two-stage training strategy, progressively adapting the model from multi-modality video generation to multi-modal controllable video synthesis with the support of X-conditioned video generation and video visual understanding.
We train the model using a learning rate of 2e-5 on 8 H100 GPUs for 40K steps. 
The model is optimized using a batch size of 8, with each training stage consisting of 20K steps. To evaluate the performance of video generation, 
we follow \cite{aetherteam2025aethergeometricawareunifiedworld} and report evaluation metrics follow VBench \cite{huang2023vbench}, a standard benchmark for video generation.

\subsection{Omni Controllable Video Generation}
We evaluate our approach against state-of-the-art methods on three tasks: text-conditioned video generation, X-conditioned video generation, and video understanding.

\paragraph{Text-conditioned video generation}
Given a text prompt, \abbname{} generates multi-modal video sequences simultaneously within a single diffusion process. 
To provide a comprehensive evaluation of our generation performance, we compare our method with the baseline video diffusion model CogVideoX \cite{yang2024cogvideox} on rgb video generation and assess the generation quality on VBench\cite{huang2023vbench} metrics. 
Note that for this comparison, we focus on the rgb modality to ensure consistency with CogVideoX, which does not support multi-modal outputs.
Table \ref{tab:vbench_prediction} presents a quantitative comparison, where our model achieves a comparable VBench metric with {CogVideoX}, demonstrating superior generation quality. 
Although our focus is on multi-modal training, the joint optimization may provide stronger regularization than using rgb alone, potentially resulting in more coherent and consistent predictions.


\paragraph{X-conditioned video generation}

\begin{figure}[t]
  \centering
  \includegraphics[width=1.0\columnwidth]{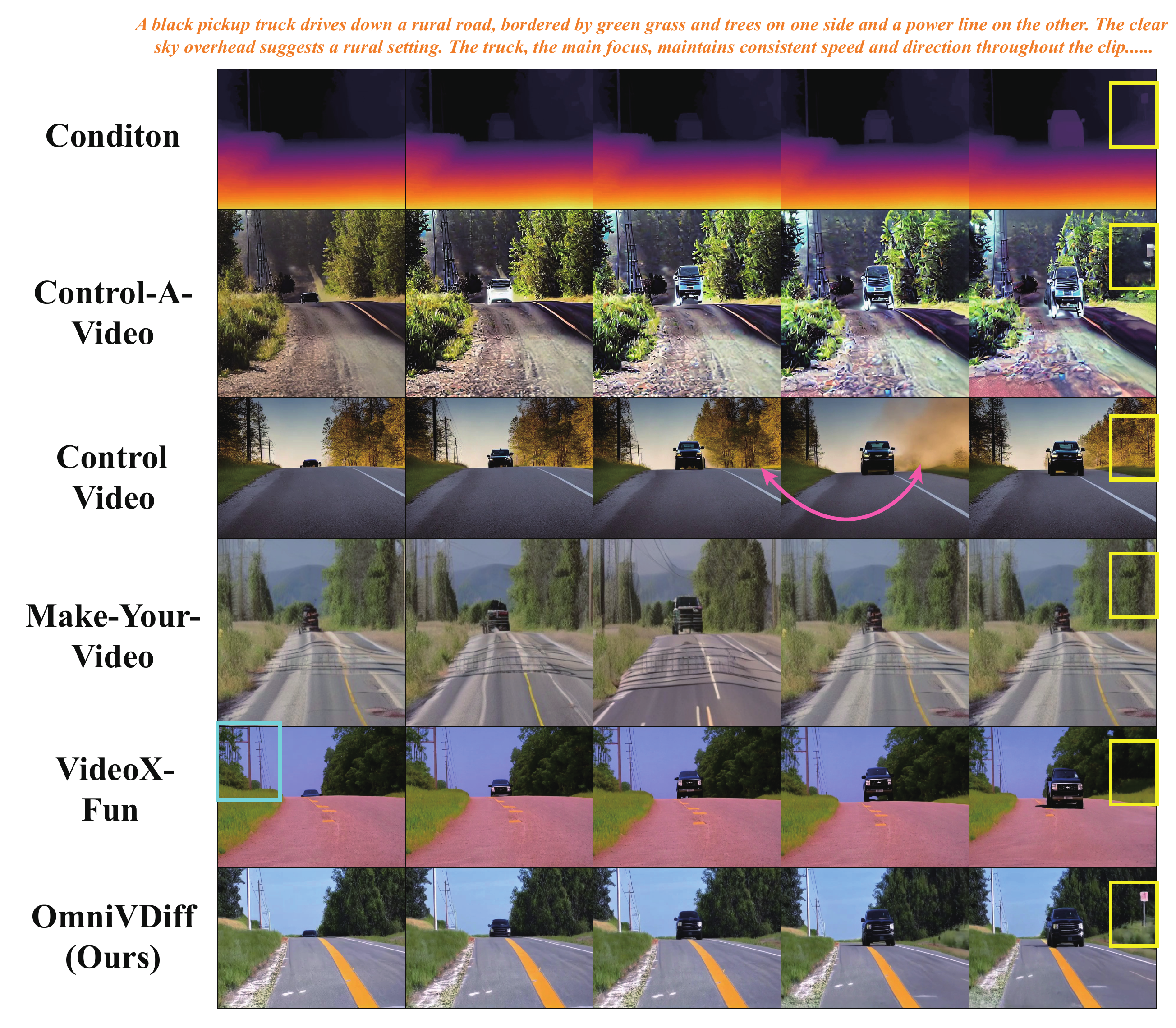}
    \caption{\textbf{Visual comparison for depth-guided video generation.} Yellow boxes highlight regions where our method better aligns with the provided depth compared to the baseline. Red arrows indicate temporal flickering, while cyan boxes denote artifacts in the rgb outputs.}

  \label{fig: Compare text depth generate}
  \vspace{-8pt}
\end{figure}

We evaluate our unified framework on X-conditioned video synthesis, comparing it with specialized baselines that leverage visual cues such as depth, canny, or segmentation.
As shown in Table~\ref{tab:comp_c2v_vbench} and Figure~\ref{fig: Compare text depth generate}, our model outperforms depth-specific baselines in depth-conditioned video generation, exhibiting superior structural fidelity and stronger alignment with the depth guidance signal.  Furthermore, Table~\ref{tab:comp_c2v_vbench} also demonstrates that our approach surpasses existing modality-specific methods in segmentation- and canny-guided synthesis. Benefiting from a unified diffusion architecture, our model enables controllable video synthesis across multiple modalities within a single cohesive framework. See the supplementary file for more details.

\begin{table*}[t]
\centering
\resizebox{\textwidth}{!}{
\begin{tabular}{lccccccc}
\toprule
 & subject consistency & b.g. consistency & motion smoothness & dynamic degree & aesthetic quality & imaging quality & weighted average \\
\midrule

w/o modality embedding & 97.11 & 95.59 & 98.97 & \underline{41.80} & 50.25 & 66.43 & \underline{71.54} \\
w/o AMCS & \underline{97.31} & \underline{96.19} & 99.01 & 33.28 & \underline{50.82} & \textbf{67.31} & 71.21 \\
w/o MSPH & 96.76 & 95.44 & \underline{99.12} & 41.41 & {50.26} & 65.81 & 71.35 \\
\abbname(Ours) & \textbf{97.78} & \textbf{96.26} & \textbf{99.21} & \textbf{49.69} & \textbf{51.47} & \underline{67.13} & \textbf{72.78} \\

\bottomrule
\end{tabular}}
\caption{\textbf{VBench metrics for the ablation study under different training settings.} For each group of metrics, the best performance is highlighted in \textbf{bold}, and the second-best is indicated with an \underline{underline}.}
\label{tab:comp_ablation}
\vspace{-8pt}
\end{table*}


\begin{figure}[t]
  \centering
  \includegraphics[width=1.0\columnwidth]{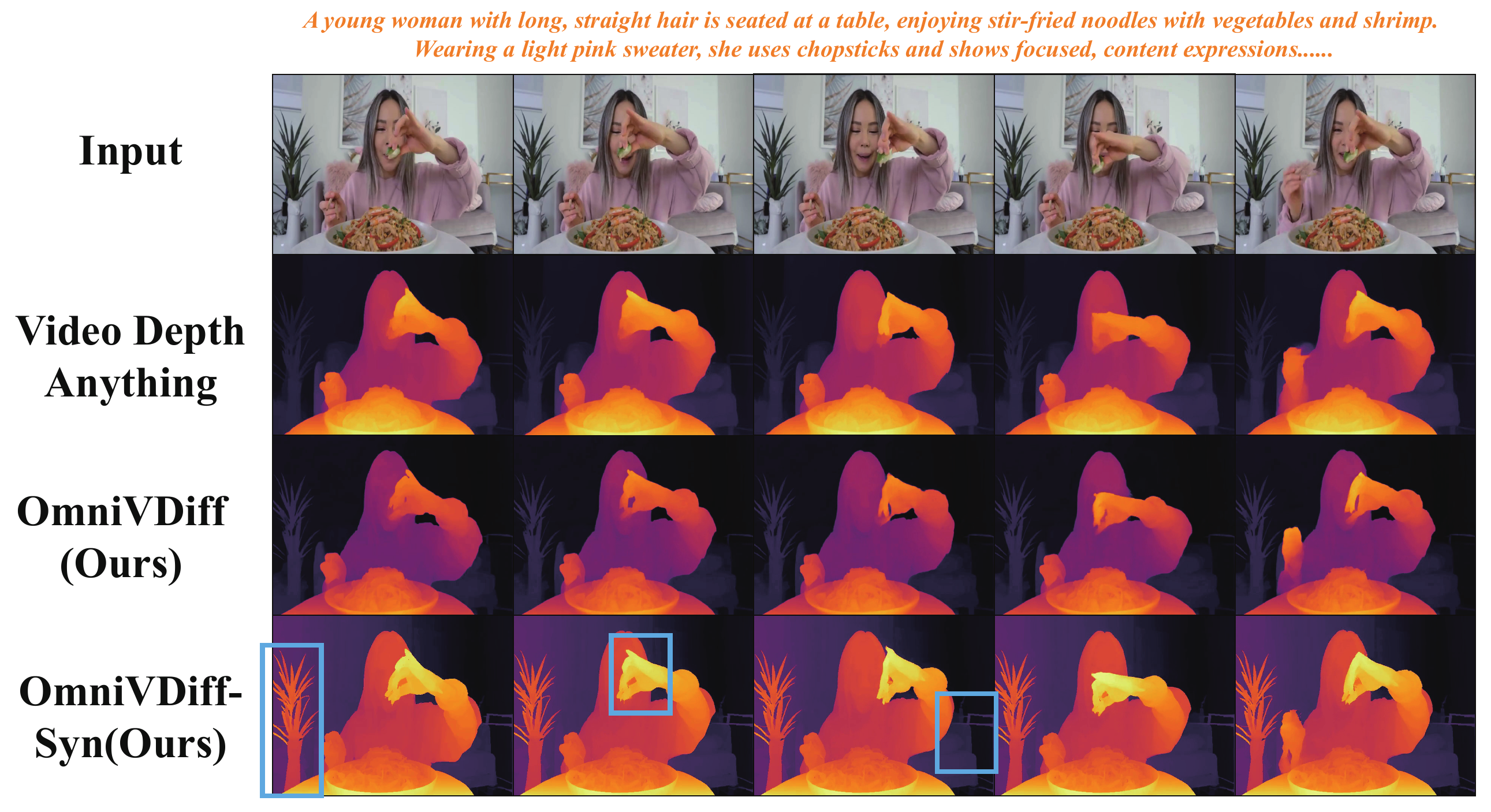}
\caption{\textbf{Qualitative comparison of video depth estimation.} 
Yellow boxes highlight areas where both \abbname{}-Syn succeed in capturing sharper details and achieving superior geometric fidelity.
}  
\label{fig:appendix_comp_understanding_depth}
\vspace{-12pt}
\end{figure}

\begin{figure}[t]
  \centering
  \includegraphics[width=1.0\columnwidth]{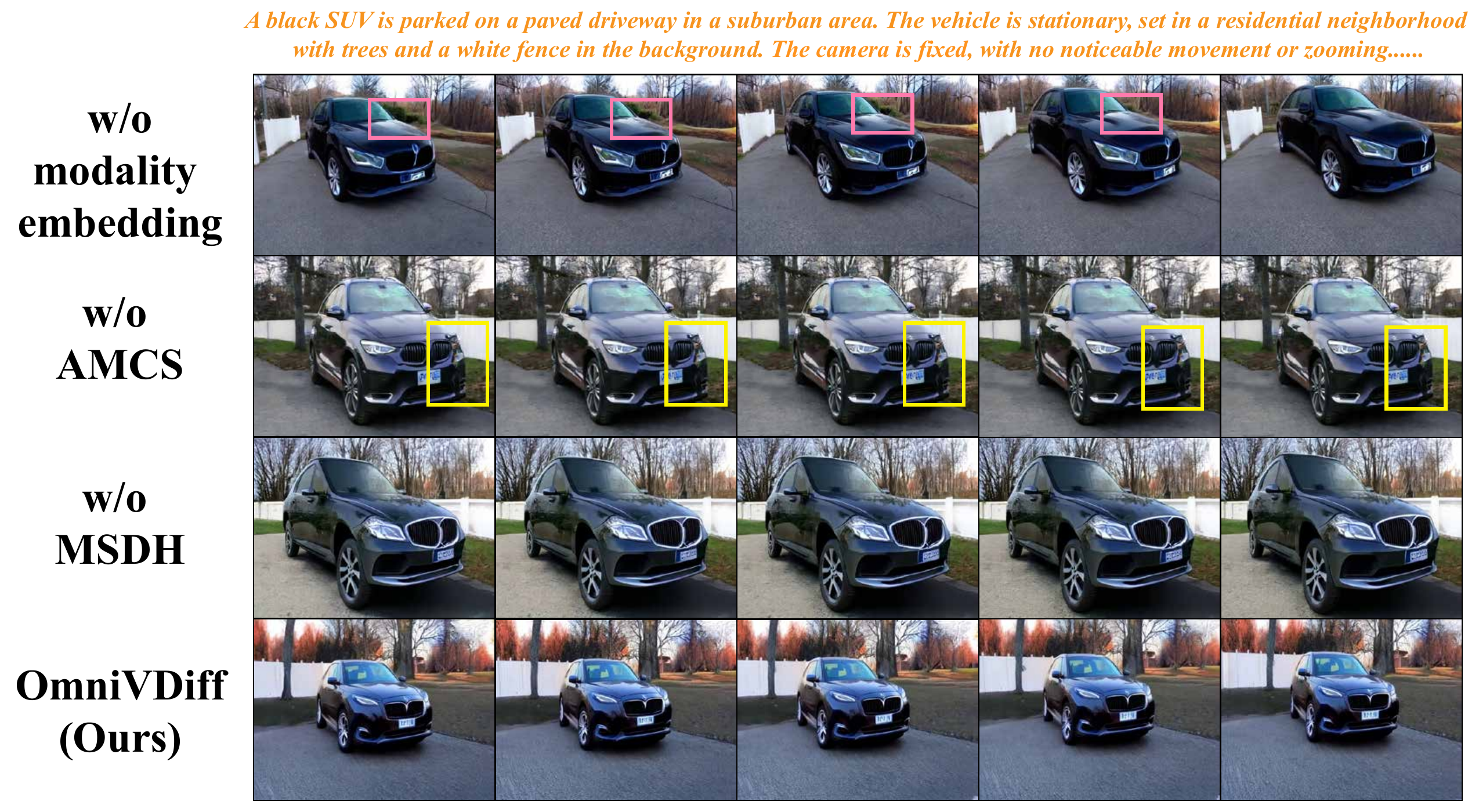}
    \caption{\textbf{Qualitative comparison of ablation variants under different training configurations.} Red boxes highlight missing rearview mirrors in the generated vehicles, while yellow boxes indicate visual artifacts.}
\label{fig:ablation experiment}
\end{figure}

\paragraph{Rgb-conditioned video understanding}

To assess video understanding capability, we compare our model against baselines specifically designed for depth and segmentation estimation.

For depth estimation, we follow the Video Depth Anything protocol~\cite{video_depth_anything} and evaluate the zero-shot performance on the ScanNet dataset~\cite{dai2017scannetrichlyannotated3dreconstructions}.
As shown in Table~\ref{tab:video_depth}, \abbname{} achieves state-of-the-art performance among all baselines, delivering results comparable to the expert model VDA-S. Notably, VDA-S serves as our teacher model and is trained with high-quality ground-truth depth supervision, while \abbname{} is trained solely with pseudo labels generated by VDA-S.

Although designed for controllable video diffusion, our model may benefit from high-quality ground-truth data for understanding tasks. We ablate this by introducing a small set of 10k synthetic samples into the training data. With this setting, \abbname{}-Syn surpasses VDA-S in accuracy and produces sharper, more precise geometric details (Figure~\ref{fig:appendix_comp_understanding_depth}). This demonstrates the model’s ability to leverage small amounts of high-quality data for significant performance gains.

Similarly, Table~\ref{tab:comparison_seg} presents quantitative comparisons on segmentation estimation, where our method achieves superior performance over baseline methods. Additional results are provided in the supplementary material.

\begin{table}[t]
\centering

\begin{tabular}{ccc}
\toprule
Method & AbsRel $\downarrow$ & $\delta_1$ $\uparrow$  \\
\midrule
DAv2-L\cite{yang2024depthv2}             & 0.150 & 0.768  \\
NVDS\cite{wang2023neural}               & 0.207 & 0.628  \\
NVDS + DAv2-L      & 0.194 & 0.658  \\
ChoronDepth \cite{shao2024learningtemporallyconsistentvideo}        & 0.199 & 0.665  \\
DepthCrafter\cite{hu2024depthcraftergeneratingconsistentlong}       & 0.169 & 0.730  \\

\bottomrule
VDA-S (e)\cite{video_depth_anything}          & \underline{0.110} & \underline{0.876}  \\
\abbname(Ours)   &  0.125 & 0.852  \\
\abbname-Syn(Ours) & \textbf{0.100} & \textbf{0.894} \\
\bottomrule
\end{tabular}
\caption{\textbf{Zero-shot video depth estimation results.} We compare our method with representative single-image  and video depth estimation models. “VDA-S(e)” denotes the expert model with a ViT-Small backbone. The \textbf{best} and \underline{second-best} results are highlighted. }
\label{tab:video_depth}
\end{table}


\begin{table}[h]
    \centering
    \resizebox{0.47\textwidth}{!}{ 
    \begin{tabular}{lcc}
        \toprule
        Method & \multicolumn{2}{c}{COCO Val 2017\cite{lin2015microsoftcococommonobjects}} \\
        \cmidrule(lr){2-3}
        & Point (Max) 1-IoU $\uparrow$ & Point (Oracle) 1-IoU $\uparrow$ \\
        \midrule
        SAM (B)\cite{kirillov2023segment}            & 52.1 & 68.2 \\
        SAM (L)\cite{kirillov2023segment}            & 55.7 & 70.5 \\
        Semantic-SAM (T)\cite{li2023semantic}   & 54.5 & 73.8 \\
        Semantic-SAM (L)(e)\cite{li2023semantic}   & \textbf{57.0} & \textbf{74.2} \\
        \abbname(ours) & \underline{56.0} & \underline{73.9}\  \\
        \bottomrule
    \end{tabular}
    } 
    \caption{\textbf{Comparison with prior methods on point-based interactions, evaluated on COCO Val2017.} “Max” selects the prediction with the highest confidence score, while “Oracle” uses the one with highest IoU against the target mask.}
    \label{tab:comparison_seg}
\end{table}

\paragraph{Ablation study}
We conduct an ablation study to assess the contributions of key design components, focusing specifically on the \emph{modality embedding}, \emph{adaptive modality control strategy (AMCS)}, and the \emph{modality-specific projection heads (MSPH)}.
As shown in Table~\ref{tab:comp_ablation} and Figure~\ref{fig:ablation experiment}, the full model consistently outperforms all ablated variants across all modalities. Introducing modality embeddings improves the model’s understanding of each modality’s role, whether as conditioning or generation input. The use of adaptive modality control facilitates flexible multi-modal control and understanding. Moreover, modality-specific projections allow the model to better capture the unique characteristics of each modality.
Together, the results confirm that these designs play a crucial role in enabling precise control and faithful synthesis in our unified diffusion framework.

\paragraph{Inference efficiency}
Our unified model offers significant efficiency advantages by supporting multi-modal video outputs within a single framework. Compared to CogVideoX, which generates only rgb videos, our model additionally produces segmentation and depth outputs with comparable inference speed and memory usage (Table~\ref{tab:memory_cost}). Moreover, unlike pipelines that rely on separate expert models for each modality—incurring substantial overhead (e.g., segmentation requires 30 seconds via separate inference)—our unified design reduces total inference time and eliminates the need to deploy multiple networks.

\subsection{Applications}
\begin{figure}
    \centering
    \begin{overpic}[width=0.9\linewidth]{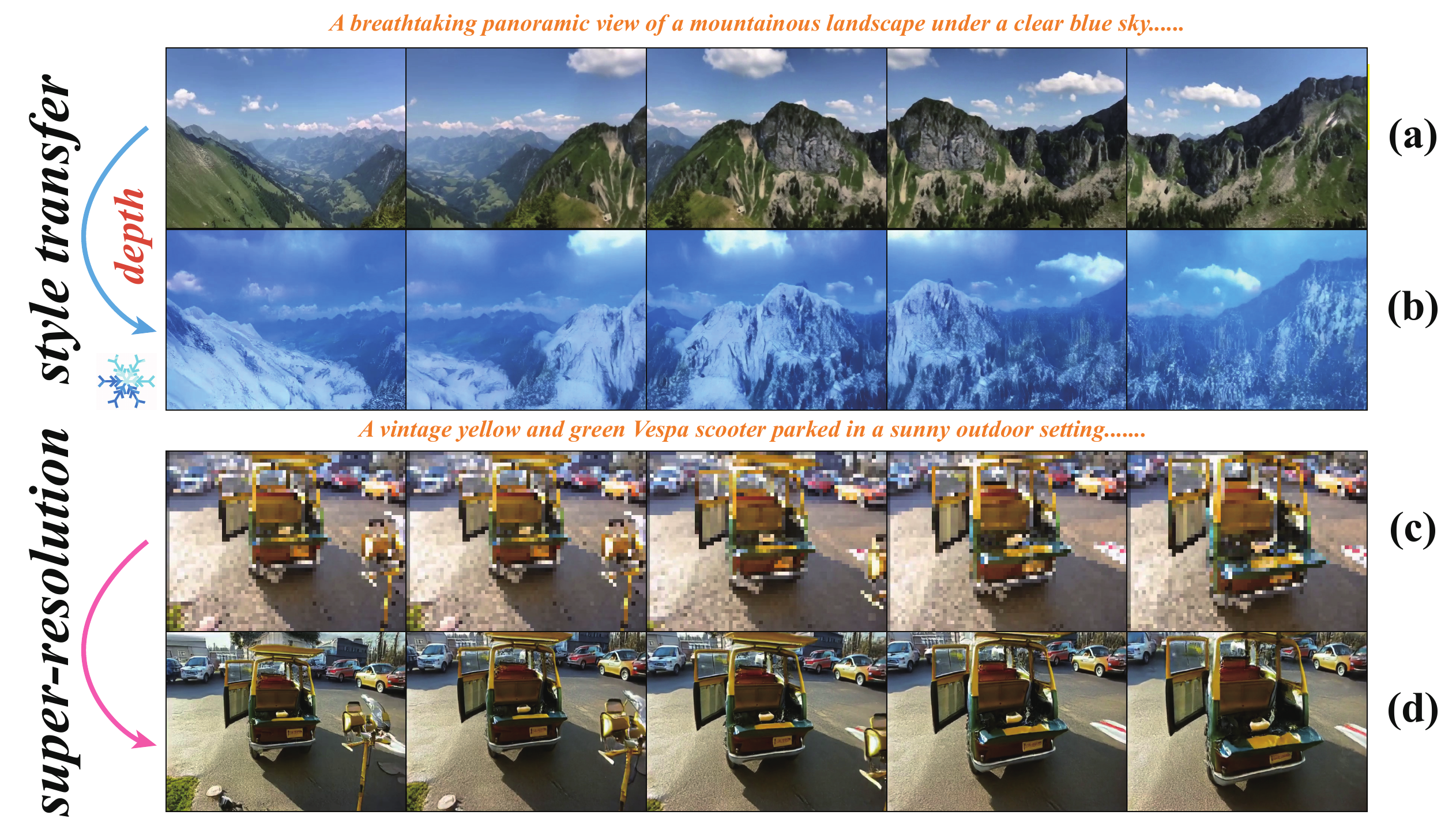}
    \end{overpic}
    \caption{
    \textbf{Applications}: 
    (a, b): Video-to-video style control. (c, d): Adapt to new tasks: video
super-resolution.
    }
    \label{fig:app_v2}
\end{figure}

Our unified model provides significant advantages in controllability and flexibility. In this section, we showcase its versatility through two representative applications:

\paragraph{Video-to-video style control}
\abbname{} can be directly applied to video-to-video style control, enabling structure-preserving video generation guided by text prompts. Given a reference video (Figure~\ref{fig:app_v2} (a)), \abbname{} first estimates depth modality as an intermediate representation, which is then used to generate diverse scene styles (Figure~\ref{fig:app_v2} (b)) (e.g., winter), while preserving the original spatial layout. Thanks to joint training, \abbname{} achieves this without relying on external depth experts, ensuring structural consistency. 
We further provide a quantitative comparison of video-to-video style control using \abbname{}’s estimated depth versus expert-provided depth, demonstrating comparable consistency and visual quality (see supplementary for details).

\paragraph{Adaptability to new modalities/tasks}
To evaluate our model’s adaptability to new modalities and applications, we conduct experiments on a representative task: video super-resolution. Specifically, we fine-tune \abbname{} for 2k steps, repurposing an existing modality slot (canny) to handle low-resolution rgb videos during training. At inference, these inputs serve as conditioning signals (Figure~\ref{fig:app_v2} (c)), enabling the model to generate high-resolution outputs (Figure~\ref{fig:app_v2} (d)), demonstrating its flexibility in handling unseen modalities with minimal adjustments.

\begin{table}[!t]
\centering
\resizebox{0.9\linewidth}{!}{
\begin{tabular}{cccc}
\hline
Methods & Paras & Time & Memory \\
\hline
Video Depth Anything     & 28.4M             & 4s  & 13.62GB  \\
Semantic-Sam \& SAM2     & 222.8 \& 38.9M    & 30s & 6.75GB   \\
CogVideoX                & 5B                & 41s & 26.48GB  \\
\hline

\abbname(Ours)        & \textbf{5B+11.8M}        & \textbf{44s} & \textbf{26.71GB}  \\

\hline
\end{tabular}
}
\caption{Comparison of Model Inference Time, Memory Usage, and Parameter Size. \abbname{} demonstrates its inference efficiency among compared models.}
\label{tab:memory_cost}
\vspace{-0.1cm}
\end{table}

\section{Conclusion}
\label{sec:conclusion}

In this paper, we present \abbname{}, a unified framework for multi-modal video generation and understanding that extends diffusion models to support text-to-video, modality-conditioned generation, and visual understanding within a single architecture. By simultaneously generating multiple modalities (i.e., rgb, depth, segmentation, and canny) and incorporating an adaptive modality control strategy, our approach flexibly handles diverse generation and conditioning scenarios. Furthermore, our unified design eliminates the need for separate expert models and sequential processing pipelines, offering a scalable and efficient solution that easily adapts to new modalities while maintaining high performance across video tasks.
Future research can explore expanding modality support, adopting more powerful pretrained models (like WAN \cite{wan2025}), and enhancing real-time efficiency, further advancing the capabilities of unified video diffusion models.

\bibliography{aaai2026}
\end{document}